\documentclass[runningheads]{llncs}
\usepackage{booktabs}
\usepackage{graphicx}
\begin{document}
\title{Synthetic DOmain-Targeted Augmentation (S-DOTA) Improves Model Generalization in Digital Pathology}
%
\titlerunning{Synthetic DOmain-Targeted Augmentation (S-DOTA)}
%

\author{Sai Chowdary Gullapally \inst{1} \and Yibo Zhang \inst{1} \and Nitin Kumar Mittal \inst{1} \and Deeksha Kartik \inst{1} \and  Sandhya Srinivasan \inst{1} \and Kevin Rose \inst{1} \and Daniel Shenker \inst{1} \and Dinkar Juyal \inst{1} \and Harshith Padigela \inst{1} \and Raymond Biju \inst{1} \and Victor Minden  \and Chirag Maheshwari  \and Marc Thibault \inst{1} \and Zvi Goldstein \inst{1} \and Luke Novak \inst{1} \and Nidhi Chandra \inst{1} \and Justin Lee \inst{1} \and Aaditya Prakash \and Chintan Shah \inst{1} \and John Abel \inst{1} \and Darren Fahy \inst{1} \and Amaro Taylor-Weiner \inst{1} \and Anand Sampat \inst{1}}
\authorrunning{S. Gullapally et al.}
%
\institute{PathAI Inc, Boston MA, USA \\ 
\email{sai.gullapally@pathai.com}}

\maketitle              
\begin{abstract}
Machine learning algorithms have the potential to improve patient outcomes in digital pathology. However, generalization of these tools is currently limited by sensitivity to variations in tissue preparation, staining procedures and scanning equipment that lead to domain shift in digitized slides. To overcome this limitation and improve model generalization, we studied the effectiveness of two Synthetic DOmain-Targeted Augmentation (S-DOTA) methods, namely CycleGAN-enabled Scanner Transform (ST) and targeted Stain Vector Augmentation (SVA), and compared them against the International Color Consortium (ICC) profile-based color calibration (ICC Cal) method and a baseline method using traditional brightness, color and noise augmentations. We evaluated the ability of these techniques to improve model generalization to various tasks and settings: four models, two model types (tissue segmentation and cell classification), two loss functions, six labs, six scanners, and three indications (hepatocellular carcinoma (HCC), nonalcoholic steatohepatitis (NASH), prostate adenocarcinoma). We compared these methods based on the macro-averaged F1 scores on in-distribution (ID) and out-of-distribution (OOD) test sets across multiple domains, and found that S-DOTA methods (i.e., ST and SVA) led to significant improvements over ICC Cal and baseline on OOD data while maintaining comparable performance on ID data. Thus, we demonstrate that S-DOTA may help address generalization due to domain shift in real world applications.
\end{abstract}
%
%

\section{Introduction}
Histopathology is the gold standard for clinical diagnosis and cancer grading. Digitization of whole-slide images (WSIs) has enabled the use of machine learning (ML) algorithms to classify diseases and quantify the biological composition of diseased tissue~\cite{Wang2016,Diao2021}. However, there are numerous sources of variations during tissue processing and digitization that lead to perceptible differences in the appearances of the resulting WSIs despite several attempts to standardize this procedure~\cite{inoue2020color}. This data variability hampers the ability of ML models to generalize to unseen domains routinely encountered in real-world settings, potentially limiting their utility.  

Domain generalization techniques aim to address this limitation of ML in digital pathology. Common methods include learning better representations by adding explicit constraints \cite{Ganin2015DomainAdversarialTO}, using contrastive learning-based losses \cite{Khosla2020SupervisedCL}, transferring target images to the source domain by using color statistics \cite{macenko2009method,Vahadane16}, using deep learning-based mapping methods \cite{Zhu2017UnpairedIT,Fick21}, calibration methods which transform both source and target images to a uniform color space at train and test time \cite{ingale2022effects}, and adding pathology-specific augmentations \cite{Hetz23,Runz21}.

In this work, we develop domain generalization techniques that aim to tackle domain shift by transforming in-distribution (ID) data to simulate out-of-distribution (OOD) data in a targeted manner.  We name this approach Synthetic DOmain-Targeted Augmentation (S-DOTA).  Specifically, we present two S-DOTA methods, (a) CycleGAN-enabled Scanner Transform (ST) and (b) physics-based, targeted Stain Vector Augmentation (SVA), and compared them against an International Color Consortium (ICC) profile-based color calibration (ICC Cal) method and baseline augmentations.  Motivated by the importance for pathology ML models to have a consistent performance across histology labs and scanners, which are major sources of real-world image variations, we quantified the relative improvements to the generalization to unseen labs and scanners achieved by these methods. The main contributions of this work are: 

\begin{itemize}
    \item We develop a generalizable approach for training a CycleGAN-based ST that transforms image patches from one scanner domain to another, which can be deployed across a wide range of unseen labs, scanners and indications. 
    \item We introduce SVA, a physics-based color transform method, which transforms image patches to diverse target stain color domains represented by stain vectors curated across a wide range of sources.
    \item We then use these S-DOTA methods as train-time augmentations to mimic data from multiple target domains of interest, and show that they improve model generalization.
    \item We compare these two S-DOTA methods against ICC Cal and baseline across multiple tissue and cell classification models with varying task complexities. We generate quantitative comparisons across ID and OOD data for liver and prostate pathology, specifically for hepatocellular carcinoma (HCC), nonalcoholic steatohepatitis (NASH) and prostate adenocarcinoma. We verify the effectiveness of these methods with Supervised Contrastive Loss (SCL) along with Cross Entropy (CE) for HCC.
\end{itemize}

\section{Methods}

\subsection{ICC Profile-based Color Calibration (ICC Cal)}

When a scanner digitizes a glass slide, it outputs an image in a specific color space. Different scanners may output images in different color spaces, i.e. the same real-world color would be represented by different digital $(R, G, B)$ values. Hence, models trained on images from one color space may be confounded by examples from a different color space. A color space is described in a standardized format known as the ICC profile. ICC Cal converts images of different color spaces to the standard RGB (sRGB) color space \cite{anderson1996proposal}. We applied this to patches for training and inference as a method for mitigation of test-time domain shift.

\subsection{Scanner Transform (ST)} CycleGANs\cite{Zhu2017UnpairedIT} have been used before to transform data between labs or scanners \cite{nerrienet2023standardized,Bouteldja22,Fick21,breen2022assessing}. Learning the transform between labs requires simultaneously learning the mapping across different scanners and different staining protocols being used at each lab, which can cause mode collapse. In addition, training one CycleGAN for every single pair of labs is impractical for real-world usage. We therefore used CycleGANs as a Scanner Transform to transform data from one scanner domain to another scanner domain as there are a tractable number of scanners. While this has been explored before for mitosis detection ~\cite{breen2022assessing,aubreville2023mitosis}, the results were inconclusive and the same data distributions were used to train both the CycleGAN and the downstream pathology model. In this work, to train the CycleGANs to only learn the transform between a given pair of scanners while being robust to inter-lab variations, we ensured each training example is a pair of unregistered image patches individually sampled from the WSIs of the same glass slide imaged by the two scanners, respectively.  40 slides stratified across labs,indications (i.e., 80 WSIs) were used to train each CycleGAN.

Once trained, we deployed the STs on source scanner data (e.g., AT2) from unseen labs and indications. We also assessed using STs on scanners which have similar color profiles to the source scanner (e.g., ScanScope, ATTurbo) and found they work comparably (see Table~\ref{tab:ST_SVA_info}).  

\subsection{Stain Vector Augmentation (SVA)} SVA decomposes stain colors of patches via linear decomposition in the absorbance space using the source patch's stain vectors estimated from the whole slide, followed by linear reconstruction of an augmented patch using target stain vectors.  Unlike previous work that either normalized (i.e., standardized) the stain color (by replacing original stain vectors with standard ones) \cite{Tellez2019QuantifyingTE} or randomly perturbed the original stain vectors to achieve augmentation \cite{MARINI2023100183}, we randomly sampled target stain vectors from a curated library of hematoxylin and eosin (H\&E) stain vectors estimated from 1631 slides, which were diversified across 21 labs, seven scanners and 45 indications (see Table~\ref{tab:ST_SVA_info}). This method allowed us to explicitly engineer the stain color distribution of the augmented patches to improve model generalization.  Stain vector estimation was performed using an approach similar to Macenko \textit{et al.} \cite{macenko2009method}.  Non-negative least squares (NNLS) was used for the decomposition.

\begin{figure}
\includegraphics[width=\textwidth]{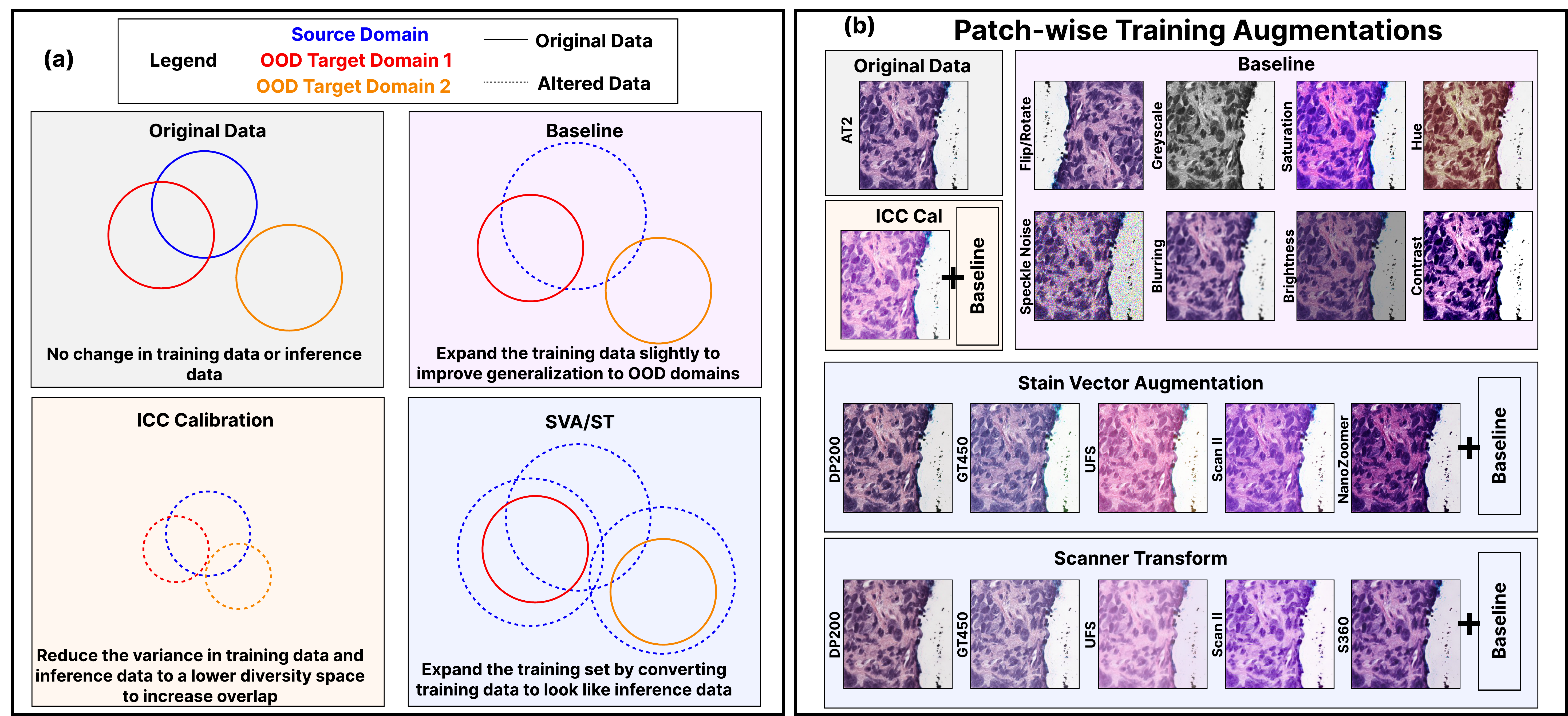}
\caption{(a) Schematic of methods and experiments on training data (blue) and OOD inference data (orange and red). (b) Example training patch from AT2 and potential augmentations during training for each experiment. SVA and ST transform the patch to multiple target domains while ICC Cal transforms the patch to the sRGB domain} \label{fig1}
\end{figure}

\begin{center}
\begin{table*}
\caption{Scanner Transform (ST) and Stain Vector Augmentation (SVA) method data table. Note that there is no overlap between the data used to develop ST and SVA in column 2 and the OOD data the pathology models were evaluated on in Table~\ref{tab:eval_data}}
    \resizebox{\textwidth}{!}{%
    \begin{tabular}{c|c|c|c|c}
    \toprule
    Technique                     & \multicolumn{1}{|c|}{What was developed?}                                    & \multicolumn{1}{|c|}{Data used for the technique}                 & \multicolumn{1}{|c}{Data on which technique was deployed}                                                                                                                                   
    \\ \midrule
             & \multicolumn{1}{|c|}{Trained CycleGANs}          & \multicolumn{1}{|c|}{CycleGAN training data}          & \multicolumn{1}{|c}{CycleGANs deployed on}              \\
             
            & \multicolumn{1}{|c|}{Scan.A$\Leftrightarrow$Scan.B}          & \multicolumn{1}{|c}{[Lab\#s list] : \#slides$[$Scan.A,Scan.B$]$}          & \multicolumn{1}{|c}{[Lab\#s list] : Source Scanners}                  \\
            
     ST         & \multicolumn{1}{|c|}{AT2$\Leftrightarrow$DP200}  & \multicolumn{1}{|c|}{$[$1,11$]$:$[$40,40$]$}          & \multicolumn{1}{|c}{$[$1,3,4,5,6,8,12$]$:$[$AT2, ATTurbo,ScanScope,NanoZoomer$]$}     \\
     
             & \multicolumn{1}{|c|}{AT2$\Leftrightarrow$GT450}          & \multicolumn{1}{|c|}{$[$1,11$]$:$[$40,40$]$
}          & \multicolumn{1}{|c}{$[$1,3,4,5,6,8,12$]$:$[$AT2,ATTurbo,ScanScope,NanoZoomer$]$}   \\
             & \multicolumn{1}{|c|}{GT450$\Leftrightarrow$DP200}          & \multicolumn{1}{|c|}{$[$1,11$]$:$[$40,40$]$}          & \multicolumn{1}{|c}{$[$1$]$:$[$GT450$]$,
}         \\
             & \multicolumn{1}{|c|}{ AT2$\Leftrightarrow$Scan II}          & \multicolumn{1}{|c|}{$[$7$]$:$[$40,40$]$}          & \multicolumn{1}{|c}{$[$1,4,8,3,6,12$]$:$[$ScanScope,AT2,ATTurbo$]$}           \\ 

& \multicolumn{1}{|c|}{ AT2$\Leftrightarrow$S360}          & \multicolumn{1}{|c|}{$[$7$]$:$[$40,40$]$}          & \multicolumn{1}{|c}{$[$1,4,8,3,6,12$]$:$[$ScanScope,AT2,ATTurbo$]$}           \\  \midrule

\               & \multicolumn{1}{|c|}{Stain Vector}  & \multicolumn{1}{|c}{Lab\#s: $[$1,2,3,4,5,7,11,13,} & \multicolumn{1}{|c}{Lab\#s:}   \\
    
&     \multicolumn{1}{|c|}{Library}  & \multicolumn{1}{|c|}{14,15,16,17,18,19,20,}  & \multicolumn{1}{|c}{$[$1,2,3,4,}   \\ 
\              & \multicolumn{1}{|c|}{stain vectors}    & \multicolumn{1}{|c|}{21,22,23,24,25,26$]$}  & \multicolumn{1}{|c}{5,6,8,12$]$} \\
      
    \  SVA    &       \multicolumn{1}{|c|}{comprised of}    & \multicolumn{1}{|c}{Scanners:$[$DP200,GT450,}          & \multicolumn{1}{|c}{Source Scanners:}    \\

    \               & \multicolumn{1}{|c|}{diverse H \& E}  & \multicolumn{1}{|c|}{AT2,ScanScope,}       & \multicolumn{1}{|c}{$[$AT2,ScanScope,}   \\ 
    
          &     \multicolumn{1}{|c|}{1631 (231/scanner)}      & \multicolumn{1}{|c}{UFS,NanoZoomer,Scan II$]$}          & \multicolumn{1}{|c}{ATTurbo,GT450,NanoZoomer$]$}           \\
     \bottomrule
    \end{tabular}%
    }
    
    \label{tab:ST_SVA_info}
    \end{table*}
    \begin{table*}
    \caption{Model details and training and validation data table for each of the pathology models used in the experiments. Each lab scanned slides using multiple scanners }
    \label{tab:model_train_data}
    \resizebox{\textwidth}{!}{%
    \begin{tabular}{c|c|c|c|c}
    \toprule
    Model                     & \multicolumn{1}{|c|}{HCC Tissue}                                    & \multicolumn{1}{|c|}{HCC Cell}                 & \multicolumn{1}{|c|}{Prostate Cell}                                                                                                    & \multicolumn{1}{|c}{NASH Cell}                                                                      
    \\ \midrule
             & \multicolumn{1}{|c|}{3:$[$ATTurbo,AT2$]$}          & \multicolumn{1}{|c|}{1:$[$GT450,AT2$]$}          & \multicolumn{1}{|c|}{1:$[$GT450,AT2,UFS$]$}          & \multicolumn{1}{|c}{1:$[$GT450,AT2$]$}        \\
            & \multicolumn{1}{|c|}{6:$[$ATTurbo,ScanScope$]$}          & \multicolumn{1}{|c|}{8:$[$AT2,ScanScope$]$}          & \multicolumn{1}{|c|}{5:$[$NanoZoomer$]$}          & \multicolumn{1}{|c}{4:$[$ScanScope$]$}        \\
    Lab\#:$[$Scanners$]$          & \multicolumn{1}{|c|}{-}          & \multicolumn{1}{|c|}{3:$[$ScanScope,AT2,ATTurbo$]$}          & \multicolumn{1}{|c|}{3:$[$ScanScope,ATTurbo,AT2$]$}          & \multicolumn{1}{|c}{6:$[$AT2$]$}        \\
             & \multicolumn{1}{|c|}{-}          & \multicolumn{1}{|c|}{4:$[$ScanScope$]$}          & \multicolumn{1}{|c|}{2:$[$AT2$]$}          & \multicolumn{1}{|c}{-}        \\
             & \multicolumn{1}{|c|}{-}          & \multicolumn{1}{|c|}{12:$[$ScanScope, AT2$]$}          & \multicolumn{1}{|c|}{4:$[$AT2$]$}          & \multicolumn{1}{|c}{-}        \\
             & \multicolumn{1}{|c|}{-}          & \multicolumn{1}{|c|}{6:$[$AT2,ScanScope$]$}          & \multicolumn{1}{|c|}{-}          & \multicolumn{1}{|c}{-}        \\ \midrule
    \# Model Params      & \multicolumn{1}{|c|}{13.2M}          & \multicolumn{1}{|c|}{8.3M}          & \multicolumn{1}{|c|}{8.3M}          & \multicolumn{1}{|c}{8.3M} \\ 
    \# Classes               & \multicolumn{1}{|c|}{4}  & \multicolumn{1}{|c|}{12}       & \multicolumn{1}{|c|}{10}            & \multicolumn{1}{|c}{12} \\ 
    Train/Val Split      & \multicolumn{1}{|c|}{75/25}          & \multicolumn{1}{|c|}{71/29}          & \multicolumn{1}{|c|}{66/34}          & \multicolumn{1}{|c}{74/26} \\ 
    \# Slides                & \multicolumn{1}{|c|}{284}  & \multicolumn{1}{|c|}{2628} & \multicolumn{1}{|c|}{939}  & \multicolumn{1}{|c}{1344} \\
    \# Annotations             & \multicolumn{1}{|c|}{6.25k}      & \multicolumn{1}{|c|}{251k}  & \multicolumn{1}{|c|}{39.8k}  & \multicolumn{1}{|c}{175k} \\ 
    \# Patches              & \multicolumn{1}{|c|}{1.48M}    & \multicolumn{1}{|c|}{1.55M}  & \multicolumn{1}{|c|}{293k}   & \multicolumn{1}{|c}{968k} \\
     \bottomrule
    \end{tabular}%
    }
    
    \end{table*}
    \begin{table*}
     \caption{Evaluation data table for each of the models used in the experiments.}
    \resizebox{\textwidth}{!}{%
    \begin{tabular}{c|c|c|c|c}
    \toprule
     Format & \multicolumn{4}{c}{Lab\#, Scanner, \# Slides, \# Annotations, \# Patches}                                                                   
    \\ \midrule
    Model  & \multicolumn{1}{|c|}{HCC Tissue} & \multicolumn{1}{|c|}{HCC Cell}  & \multicolumn{1}{|c|}{Prostate Cell} & \multicolumn{1}{|c}{NASH Cell}                                                                      \\ \midrule
    OOD 1          & \multicolumn{1}{|c|}{7,Scan II,79,1.58k,74.5k}   & \multicolumn{1}{|c|}{7,Scan II,200,20.2k,182k}          & \multicolumn{1}{|c|}{1,DP200,100,10.2k,102k}          & \multicolumn{1}{|c}{4,AT2,20,2.05k,11.1k} \\ 
    OOD 2              & \multicolumn{1}{|c|}{7,DP200,162,3.30k,118k}    & \multicolumn{1}{|c|}{7,DP200,200,20.7k,207k} & \multicolumn{1}{|c|}{-}  & \multicolumn{1}{|c}{11,AT2,20,1.87k,10.3k} \\
    OOD 3                & \multicolumn{1}{|c|}{7,GT450,119,2.02k,96.6k}    & \multicolumn{1}{|c|}{7,GT450,201,20.5k,205k}  & \multicolumn{1}{|c|}{-} & \multicolumn{1}{|c}{11,DP200,20,1.85k,9.78k} \\ 
    OOD 4                  & \multicolumn{1}{|c|}{-} & \multicolumn{1}{|c|}{-}   & \multicolumn{1}{|c|}{-} & \multicolumn{1}{|c}{11,GT450,19,1.78k,10.8k} \\ 
    
    \midrule  \midrule
                    & \multicolumn{1}{|c|}{3,ScanScope,69,2.06k,82.2k}    & \multicolumn{1}{|c|}{3,ScanScope,70,7.03k,55.4k}         & \multicolumn{1}{|c|}{3,ScanScope,20,2.88k,28.8k}           & \multicolumn{1}{|c}{4,ScanScope,13,2.95k,76.7k}        \\ 
                  & \multicolumn{1}{|c|}{3,AT2,1,40,1.43k}     & \multicolumn{1}{|c|}{3,AT2,1,115,770}           & \multicolumn{1}{|c|}{3,ATTurbo,3,20,200}          & \multicolumn{1}{|c}{1,AT2,91,9.06k,241k}        \\ 
                     & \multicolumn{1}{|c|}{-}          & \multicolumn{1}{|c|}{-}       & \multicolumn{1}{|c|}{3,AT2,1,11,110}      & \multicolumn{1}{|c}{-}        \\
      ID          & \multicolumn{1}{|c|}{-}          & \multicolumn{1}{|c|}{-}    & \multicolumn{1}{|c|}{3,GT450,3,28,280}        & \multicolumn{1}{|c}{-}        \\
                 & \multicolumn{1}{|c|}{-}          & \multicolumn{1}{|c|}{-}      & \multicolumn{1}{|c|}{4,AT2,48,2.44k,24.4k}     & \multicolumn{1}{|c}{-}        \\
              & \multicolumn{1}{|c|}{-}          & \multicolumn{1}{|c|}{-}     & \multicolumn{1}{|c|}{1,GT450,133,5.48k,54.8k}         & \multicolumn{1}{|c}{-}        \\
                & \multicolumn{1}{|c|}{-}          & \multicolumn{1}{|c|}{-}      & \multicolumn{1}{|c|}{1,AT2,80,3.25k,32.5k}      & \multicolumn{1}{|c}{-}        \\
                  & \multicolumn{1}{|c|}{-}          & \multicolumn{1}{|c|}{-}       & \multicolumn{1}{|c|}{2,AT2,4,161,1.61k}   & \multicolumn{1}{|c}{-}        \\
     \bottomrule
    \end{tabular}%
    }
    
    \label{tab:eval_data}
    \end{table*}
    
    \end{center}

\section{Experiments and Results}

\subsection{Data}
We used hematoxylin and eosin (H\&E) WSIs from 26 different labs (referenced herein as Labs 1-26, this includes one lab from TCGA~\cite{weinstein2013cancer}) digitized across 9 scanners (Aperio AT2, Aperio AT Turbo, Aperio GT450, Aperio ScanScope V1, Hamamatsu NanoZoomer, Hamamatsu S360, Philips UFS, Ventana DP200 and 3DHistech Pannoramic Scan II). We ensured no overlap amongst the cohorts.\\ \textbf{Data for developing ST and SVA}Five STs were trained with data from three labs, three indications and five scanners. 1631 slides across 21 labs, seven scanners and 45 indications were curated for the stain vector library (see Table~\ref{tab:ST_SVA_info}). \\ \textbf{Training Data for Pathology Models} Pathology models were trained with data from HCC, NASH and Prostate Cancer across eight labs and five scanners (see Table~\ref{tab:model_train_data}). ST and SVA were deployed on this data for S-DOTA.\\\textbf{Evaluation Data for Pathology Models} Pathology models are evaluated with data from HCC, NASH and Prostate Cancer across six labs and six scanners (see Table~\ref{tab:eval_data}). 
\subsection{Setup} 
 \textbf{Pathology Models}
We trained ResNet-based~\cite{DBLP:journals/corr/HeZRS15} pathology models (see Table~\ref{tab:model_train_data}) as the baselines for each experiment. All the baseline hyperparameters (patch size, learning rate, batch size, architecture) and baseline augmentations (flip, crop, rotate, grayscale, noise, hue, saturation, contrast, brightness) were tuned for optimal performance and kept constant when applied in subsequent experiments. The techniques below are applied as additional augmentations (see Figure~\ref{fig1}(b)). In total we trained 20 pathology models across 5 experiments.\\ \textbf{ICC Profile-based Color Calibration} 
We obtained ICC profiles of the WSIs from the image metadata as produced by the scanner. These profiles were used to transform training data patches into the sRGB color space before the baseline augmentations are applied. During evaluation, test data patches are also transformed into the sRGB color space.\\\textbf{SVA and ST}
During training of the pathology models, we applied ST and SVA as additional train-time augmentations to the training data patches. Depending on the technique, we either apply a pre-trained CycleGAN (ST) or a stain vector transformation (SVA). We transform ID patches to all possible OOD scanners such that the group of patches passed to the model are approximately equally distributed across scanners. Then we applied baseline augmentations on top of this (see Fig.~\ref{fig1}(b)). We apply no transformations to patches at inference-time and deploy trained pathology models on real ID and OOD test data.

\subsection{Results and Discussion}
We compared macro-averaged F1 scores (see Tables ~\ref{tab:hcc_tissue_f1},~\ref{tab:hcc_cell_f1},~\ref{tab:prostate_f1},~\ref{tab:nash_f1}) for all methods across OOD and ID data (see Table~\ref{tab:eval_data}). The confidence intervals were obtained by bootstrapping over test-set tissue region and cell point annotations (10 rounds) and 5th, 50th and 95th metric percentiles are shown.\\\textbf{HCC Tissue Model} (Table~\ref{tab:hcc_tissue_f1}). SVA and ST strictly outperformed ICC Cal, baseline on all OOD data. Scan II improved the most. We hypothesize this is likely because baseline had a higher generalization gap on Scan II.  All methods are better than baseline on ID data. Overall ST and SVA are comparable and both are better than ICC Cal and baseline. \\ \textbf{HCC Cell Model} (Table~\ref{tab:hcc_cell_f1}) SVA and ST strictly outperformed ICC Cal, baseline on all OOD data. Improvement on GT450 is lower compared to Scan II. We hypothesize this is because the GT450 data from Lab 1 used in training the baseline (see Table~\ref{tab:model_train_data}) likely caused it to have a much smaller generalization gap on GT450 OOD data from Lab 7. All methods are better than baseline on ID data. Overall ST and SVA are comparable and both are better than ICC Cal and baseline. \\ \textbf{Prostate Cell Model} (Table~\ref{tab:prostate_f1}) SVA and ST strictly outperformed ICC Cal, baseline on all OOD data. On ID data, while all of the methods are inferior to baseline, this drop in performance is much smaller than the gain in performance on OOD data. We hypothesize the baseline overfits onto ID data thus resulting in a slightly higher performance on ID data. Therefore overall ST and SVA are comparable and both are better than ICC Cal and baseline. \\ \textbf{NASH Cell Model} (Table~\ref{tab:nash_f1}) ST, SVA and ICC Cal are all better than baseline on OOD and ID data. ST is the most robust across different labs and scanners and performs best overall.\\ \textbf{Supervised Contrastive Loss (SCL)} (Table~\ref{tab:hcc_cell_scl_f1})
Given the popularity of SCL, we ran ablation experiments on the HCC Cell model to study the effectiveness of these approaches with SCL. ST and SVA strictly outperform ICC Cal and baseline on Scan II and GT450, and are comparable to ICC Cal and baseline on DP200. On ID data, SVA and ST are better while ICC Cal is inferior to baseline. Overall we see ST and SVA are just as effective and better than ICC Cal when used with SCL, as with cross entropy loss.\\ \textbf{Discussion} ST and SVA perform comparably across different tasks, and both are more effective than ICC Cal likely because multiple STs and a diverse stain vector library enabled generation of a diverse set of synthetic patches that accurately mimic the OOD domain, whereas ICC Cal does not account for variations in the tissue preparation process. Moreover, ICC Cal, unlike ST and SVA, requires transforming data at inference-time which may cause regulatory challenges. To transform the source patch to $n$ domains using SVA, we only need $n$ stain vectors (computed from $n$ slides from the domains of interest); by contrast, for ST we need $n$ trained CycleGANs which requires more data and compute time.  Therefore, in practice, SVA can often generate more diverse data than ST. Considering the ease of use and performance, we recommend SVA, ST, and ICC Cal in that order for digital pathology models applied to H\&E WSIs.

\begin{center}
\begin{table*}
\caption{Comparison of macro F1 score x 100 for \textbf{HCC Tissue} model on OOD data-sets (\#Lab, Scanner) and ID datasets specified in Table~\ref{tab:eval_data}.}
\resizebox{\textwidth}{!}{%
\begin{tabular}{ccccc}
\toprule
Dataset & \multicolumn{1}{|c|}{OOD 1 (7,Scan II)}                                                      & \multicolumn{1}{|c|}{OOD 2 (7,DP200)}                                                                 & \multicolumn{1}{|c|}{OOD 3 (7,GT450)}                                                      & \multicolumn{1}{||c}{ID}                                                                               \\ \cmidrule(lr){1-1} \cmidrule(lr){2-2} \cmidrule(lr){3-3} \cmidrule(lr){4-4} \cmidrule(lr){5-5}  
Method & \multicolumn{1}{|c|}{50\%$[$5\%,95\%$]$} & \multicolumn{1}{|c|}{50\%$[$5\%,95\%$]$} & \multicolumn{1}{|c|}{50\%$[$5\%,95\%$]$} & \multicolumn{1}{||c}{50\%$[$5\%,95\%$]$}                                                                \\ \midrule
Baseline          & \multicolumn{1}{|c|}{77.79$[$76.59,79.11$]$}               & \multicolumn{1}{|c|}{87.66$[$86.73,88.31$]$} & \multicolumn{1}{|c|}{88.96$[$87.99,89.68$]$}   & \multicolumn{1}{||c}{79.52$[$77.69,80.49$]$}    \\ \midrule
ICC Cal          & \multicolumn{1}{|c|}{84.66$[$83.35,86.54$]$}              & \multicolumn{1}{|c|}{87.85$[$86.67,88.53$]$}        & \multicolumn{1}{|c|}{92.51$[$91.80,93.20$]$}  & \multicolumn{1}{||c}{81.09$[$80.60,82.37$]$}   \\ 
ST        & \multicolumn{1}{|c|}{88.31$[$87.80,89.60$]$}    & \multicolumn{1}{|c|}{\textbf{89.74$[$89.16,90.10$]$}}       & \multicolumn{1}{|c|}{93.61$[$93.25,93.95$]$}   & \multicolumn{1}{||c}{80.71$[$80.05,81.47$]$}     \\ 
SVA          & \multicolumn{1}{|c|}{\textbf{88.64$[$87.48,89.28$]$}}  & \multicolumn{1}{|c|}{89.20$[$88.58,89.92$]$}  & \multicolumn{1}{|c|}{\textbf{93.73$[$93.50,94.04$]$}} & \multicolumn{1}{||c}{\textbf{82.07$[$80.82,83.59$]$}} \\ 
 \bottomrule
\end{tabular}%
}
\label{tab:hcc_tissue_f1}
\end{table*}
\begin{table*}
\caption{Comparison of macro F1 score x 100 for \textbf{HCC Cell} model on OOD datasets (\#Lab, Scanner) and ID datasets specified in Table ~\ref{tab:eval_data}}
\resizebox{\textwidth}{!}{%
\begin{tabular}{ccccc}
\toprule
Dataset & \multicolumn{1}{|c|}{OOD 1 (7,Scan II)}                                                      & \multicolumn{1}{|c|}{OOD 2 (7,DP200)}                                                                 & \multicolumn{1}{|c|}{OOD 3 (7,GT450)}                                                      & \multicolumn{1}{||c}{ID}                                                                               \\ \cmidrule(lr){1-1} \cmidrule(lr){2-2} \cmidrule(lr){3-3} \cmidrule(lr){4-4} \cmidrule(lr){5-5}  
Method & \multicolumn{1}{|c|}{50\%$[$5\%,95\%$]$} & \multicolumn{1}{|c|}{50\%$[$5\%,95\%$]$} & \multicolumn{1}{|c|}{50\%$[$5\%,95\%$]$} & \multicolumn{1}{||c}{50\%$[$5\%,95\%$]$}                                                                \\ \midrule 
Baseline          & \multicolumn{1}{|c|}{55.01$[$54.84,55.52$]$}               & \multicolumn{1}{|c|}{65.41$[$64.79,65.91$]$} & \multicolumn{1}{|c|}{63.09$[$62.74,63.49$]$}   & \multicolumn{1}{||c}{66.83$[$66.09,67.36$]$}    \\ \midrule
ICC Cal          & \multicolumn{1}{|c|}{50.56$[$50.24,51.03$]$}              & \multicolumn{1}{|c|}{66.61$[$65.94,67.44$]$}        & \multicolumn{1}{|c|}{61.01$[$60.47,61.71$]$}  & \multicolumn{1}{||c}{67.33$[$65.76,67.97$]$}   \\ 
ST        & \multicolumn{1}{|c|}{\textbf{64.05$[$63.34,64.70$]$}}    & \multicolumn{1}{|c|}{70.24$[$69.69,70.93$]$}       & \multicolumn{1}{|c|}{65.03$[$64.55,65.74$]$}   & \multicolumn{1}{||c}{\textbf{68.31$[$67.76,69.30$]$}}     \\ 
SVA          & \multicolumn{1}{|c|}{59.96$[$59.60,60.25$]$}  & \multicolumn{1}{|c|}{\textbf{72.76$[$72.38,73.39$]$}}  & \multicolumn{1}{|c|}{\textbf{65.66$[$64.99,66.14$]$}} & \multicolumn{1}{||c}{67.10$[$66.53,67.90$]$} \\ 
 \bottomrule
\end{tabular}
}
\label{tab:hcc_cell_f1}
\end{table*}
\begin{table*}
\caption{Comparison of macro F1 score x 100 for \textbf{Prostate Cell} model on OOD datasets (\#Lab, Scanner) and ID datasets specified in Table~\ref{tab:eval_data}}
\begin{center}
\resizebox{6cm}{!}{%
\begin{tabular}{c|c|c|}
\toprule
Dataset & \multicolumn{1}{|c|}{OOD 1 (1,DP200)}                                                      & \multicolumn{1}{||c}{ID}                                                                               \\ \cmidrule(lr){1-1} \cmidrule(lr){2-2} \cmidrule(lr){3-3} 
Method & \multicolumn{1}{|c|}{50\%$[$5\%,95\%$]$} & \multicolumn{1}{||c}{50\%$[$5\%,95\%$]$}                                                                 \\ \midrule
Baseline          & \multicolumn{1}{|c|}{51.54$[$50.38,51.87$]$}               & \multicolumn{1}{||c}{\textbf{64.83$[$63.92,65.83$]$}}  \\ \midrule
ICC Cal          & \multicolumn{1}{|c|}{56.95$[$55.97,57.45$]$}              & \multicolumn{1}{||c}{62.14$[$61.72,63.19$]$}        \\ 
ST          & \multicolumn{1}{|c|}{\textbf{61.88$[$61.58,62.36$]$}}              & \multicolumn{1}{||c}{64.29$[$64.01,64.64$]$}        \\ 
SVA          & \multicolumn{1}{|c|}{60.98$[$60.30,61.61$]$}  & \multicolumn{1}{||c}{61.48$[$60.68,62.42$]$} \\ 
 \bottomrule
\end{tabular}
}
\end{center}

\label{tab:prostate_f1}
\end{table*}
\begin{table*}
\caption{Comparison of macro F1 score x 100 for \textbf{NASH Cell} model on OOD datasets (\#Lab, Scanner) and ID datasets specified in Table~\ref{tab:eval_data}}
    \resizebox{\textwidth}{!}{%
    \begin{tabular}{cccccc}
    \toprule
    Dataset & \multicolumn{1}{|c|}{OOD 1 (4,AT2)}                                                      & \multicolumn{1}{|c|}{OOD 2 (11,AT2)}                               & \multicolumn{1}{|c|}{OOD 3 (11,DP200)}                                  & \multicolumn{1}{|c|}{OOD 4 (11,GT450)}                                                      & \multicolumn{1}{||c}{ID}                                                                               \\ \cmidrule(lr){1-1} \cmidrule(lr){2-2} \cmidrule(lr){3-3} \cmidrule(lr){4-4} \cmidrule(lr){5-5} \cmidrule(lr){6-6} 
    Method & \multicolumn{1}{|c|}{50\%$[$5\%,95\%$]$} & \multicolumn{1}{|c|}{50\%$[$5\%,95\%$]$} & \multicolumn{1}{|c|}{50\%$[$5\%,95\%$]$} & \multicolumn{1}{|c|}{50\%$[$5\%,95\%$]$} & \multicolumn{1}{||c}{50\%$[$5\%,95\%$]$}                                                                \\ \midrule
    Baseline          & \multicolumn{1}{|c|}{76.81$[$74.57,78.27$]$}               & \multicolumn{1}{|c|}{75.01$[$72.96,76.78$]$} & \multicolumn{1}{|c|}{81.19$[$78.19,83.14$]$} & \multicolumn{1}{|c|}{76.11$[$73.76,77.57$]$}   & \multicolumn{1}{||c}{50.58$[$49.84,51.62$]$}    \\ \midrule
    ICC Cal          & \multicolumn{1}{|c|}{74.93$[$72.53,75.81$]$}              & \multicolumn{1}{|c|}{78.75$[$76.29,80.47$]$}   & \multicolumn{1}{|c|}{85.07$[$83.56,86.23$]$}     & \multicolumn{1}{|c|}{78.28$[$76.36,79.67$]$}  & \multicolumn{1}{||c}{59.47$[$58.44,59.95$]$}   \\ 
    ST        & \multicolumn{1}{|c|}{\textbf{79.38$[$76.64,80.61$]$}}    & \multicolumn{1}{|c|}{\textbf{80.54$[$78.39,82.35$]$}}   & \multicolumn{1}{|c|}{\textbf{85.92$[$84.01,88.41$]$}}    & \multicolumn{1}{|c|}{\textbf{82.040$[$79.76,83.75$]$}}   & \multicolumn{1}{||c}{\textbf{60.22$[$59.23,61.40$]$}}     \\ 
    SVA          & \multicolumn{1}{|c|}{76.60$[$74.91,80.70$]$}  & \multicolumn{1}{|c|}{78.39$[$76.64,79.22$]$} & \multicolumn{1}{|c|}{85.11$[$82.51,86.60$]$} & \multicolumn{1}{|c|}{79.09$[$76.69,81.04$]$} & \multicolumn{1}{||c}{56.20$[$55.06,56.62$]$} \\ 
     \bottomrule
    \end{tabular}
    }
    \label{tab:nash_f1}
    \end{table*}
    \begin{table*}
    \caption{Comparison of macro F1 score x 100 for \textbf{HCC Cell} model trained with \textbf{SCL} loss on OOD datasets and ID datasets specified in Table~\ref{tab:eval_data}}
\resizebox{\textwidth}{!}{%
\begin{tabular}{ccccc}
\toprule
Dataset & \multicolumn{1}{|c|}{OOD 1 (7,Scan II)}                                                      & \multicolumn{1}{|c|}{OOD 2 (7,DP200)}                                                                 & \multicolumn{1}{|c|}{OOD 3 (7,GT450)}                                                      & \multicolumn{1}{||c}{ID}                                                                               \\ \cmidrule(lr){1-1} \cmidrule(lr){2-2} \cmidrule(lr){3-3} \cmidrule(lr){4-4} \cmidrule(lr){5-5}  
Method & \multicolumn{1}{|c|}{50\%$[$5\%,95\%$]$} & \multicolumn{1}{|c|}{50\%$[$5\%,95\%$]$} & \multicolumn{1}{|c|}{50\%$[$5\%,95\%$]$} & \multicolumn{1}{||c}{50\%$[$5\%,95\%$]$}                                                                \\ \midrule
Baseline+SCL          & \multicolumn{1}{|c|}{46.23$[$45.68,47.12$]$}               & \multicolumn{1}{|c|}{\textbf{69.53$[$68.79,69.80$]$}} & \multicolumn{1}{|c|}{60.65$[60.37,	61.63]$}   & \multicolumn{1}{||c}{65.87$[$64.82,66.69$]$}    \\ \midrule
ICC Cal+SCL          & \multicolumn{1}{|c|}{41.26$[40.69,41.43]$}              & \multicolumn{1}{|c|}{68.42$[$68.01,68.77$]$}        & \multicolumn{1}{|c|}{53.44	$[$53.00,54.00$]$}  & \multicolumn{1}{||c}{64.41$[$63.35,65.34$]$}   \\ 
ST+SCL       & \multicolumn{1}{|c|}{\textbf{60.94$[$60.19,61.91$]$}}    & \multicolumn{1}{|c|}{69.17$[$68.74,69.70$]$}       & \multicolumn{1}{|c|}{62.24	$[$61.75,62.91$]$}   & \multicolumn{1}{||c}{67.41$[$66.48,68.14$]$}     \\ 
SVA+SCL          & \multicolumn{1}{|c|}{60.00$[$59.77,60.75$]$}  & \multicolumn{1}{|c|}{68.75$[$67.99,69.12$]$}  & \multicolumn{1}{|c|}{\textbf{62.79$[$61.97,63.33$]$}} & \multicolumn{1}{||c}{\textbf{68.22$[$67.26,69.29$]$}} \\ 
 \bottomrule
\end{tabular}
}
\label{tab:hcc_cell_scl_f1}
\end{table*}
\end{center}

\section{Conclusion}
 In this work, we present S-DOTA methods (ST and SVA) for improving the domain generalization of pathology ML models. We compared ST and SVA against ICC Cal and baseline augmentations and showed that ST and SVA consistently outperformed ICC Cal and baseline across three different indications, two loss functions, and across cell and tissue classification models. Therefore, we conclude that S-DOTA reliably improved model generalization in real-world-like settings. Future work includes expansion to more model types (e.g. end-to-end models), non-H\&E stains (e.g. immunohistochemistry) and more indications to further test the effectiveness of these methods. \\\\\textbf{Acknowledgements.} We thank the following folks for helpful guidance, discussions and related experiments that informed our work: Tan Nguyen, Jun Zhang, Kishalve Pethia, Julia Varao, Archit Khosla, Chintan Parmar,  Pratik Mistry, Michael Nercessian, George Hu, Srujan Vajram, Adam Stanford-Moore, Lara Murray, Janani Iyer,  Brian Baker, Jimish Mehta, Sergine Brutus, Ben Glass, Chris Kirby, Michael G. Drage, Ryan Cabeen, Ken Leidal.

%
%
\bibliographystyle{splncs04}
\bibliography{references}

%
\title{Supplementary Material for Synthetic DOmain-Targeted Augmentation (S-DOTA) Improves Model Generalization in Digital Pathology}
\titlerunning{Synthetic DOmain-Targeted Augmentation (S-DOTA)}
\author{}
\institute{}
\maketitle

\begin{center}
\begin{table*}
\caption{Macro F1 score x 100 for \textbf{HCC Cell} model on OOD and ID datasets (\#Lab, Scanner) \textbf{when we do not use synthetic Scan II data for ST, SVA to transform patches while training the pathology model}. ST, SVA models still generalize better than Baseline likely due to high variance in data vs. Baseline, ICC Cal models (via S-DOTA) even though the variance may not directly represent what is seen in the OOD evaluation set.}
\resizebox{\textwidth}{!}{%
\begin{tabular}{ccccc}
\toprule
Dataset & \multicolumn{1}{|c|}{OOD 1 (7,Scan II)}                                                      & \multicolumn{1}{|c|}{OOD 2 (7,DP200)}                                                                 & \multicolumn{1}{|c|}{OOD 3 (7,GT450)}                                                      & \multicolumn{1}{||c}{ID}                                                                               \\ \cmidrule(lr){1-1} \cmidrule(lr){2-2} \cmidrule(lr){3-3} \cmidrule(lr){4-4} \cmidrule(lr){5-5}  
Method & \multicolumn{1}{|c|}{50\%$[$5\%,95\%$]$} & \multicolumn{1}{|c|}{50\%$[$5\%,95\%$]$} & \multicolumn{1}{|c|}{50\%$[$5\%,95\%$]$} & \multicolumn{1}{||c}{50\%$[$5\%,95\%$]$}                                                                \\ \midrule 
Baseline          & \multicolumn{1}{|c|}{55.01$[$54.84,55.52$]$}               & \multicolumn{1}{|c|}{65.41$[$64.79,65.91$]$} & \multicolumn{1}{|c|}{63.09$[$62.74,63.49$]$}   & \multicolumn{1}{||c}{66.83$[$66.09,67.36$]$}    \\ \midrule
ICC Cal          & \multicolumn{1}{|c|}{50.56$[$50.24,51.03$]$}              & \multicolumn{1}{|c|}{66.61$[$65.94,67.44$]$}        & \multicolumn{1}{|c|}{61.01$[$60.47,61.71$]$}  & \multicolumn{1}{||c}{67.33$[$65.76,67.97$]$}   \\ 
ST        & \multicolumn{1}{|c|}{58.82$[$58.46,59.42$]$}    & \multicolumn{1}{|c|}{70.1$[$69.76,70.62$]$}       & \multicolumn{1}{|c|}{66.4$[$65.79,66.75$]$}   & \multicolumn{1}{||c}{\textbf{68.72$[$67.88,69.96$]$}}     \\ 
SVA          & \multicolumn{1}{|c|}{\textbf{59.38$[$58.91,60.21$]$}}  & \multicolumn{1}{|c|}{\textbf{72.06$[$71.51,72.47$]$}}  & \multicolumn{1}{|c|}{\textbf{64.56$[$63.86,65.27$]$}} & \multicolumn{1}{||c}{67.5$[$66.23,68.17$]$} \\ 
 \bottomrule
\end{tabular}
}
\label{tab:hcc_cell_f1}
\end{table*}

\begin{table*}
\caption{Details of the synthetic training data created with S-DOTA using ST in each of the experiments. Appropriate STs were chosen to synthesize this new data. Note that STs trained for AT2 were also deployed on ATTurbo, ScanScope and NanoZoomer data due to the similarity in their color appearance. Both original and synthetic data is used to train the pathology models.}
\label{tab:train_data}

\resizebox{\textwidth}{!}{%
\begin{tabular}{c|c|c}
\toprule
Format                                &   \multicolumn{1}{c}{Original Scanner:$[$Labs$]$}    & \multicolumn{1}{c}{Original Scanner $\rightarrow$ (Synthetic Scanners)} \\ \midrule

HCC Tissue      & \multicolumn{1}{c}{AT2:$[$3$]$}   & \multicolumn{1}{c}{AT2 $\rightarrow$ (DP200, GT450, Scan II, S360)}  \\ 

& \multicolumn{1}{c}{ATTurbo:$[$3,6$]$}  & \multicolumn{1}{c}{ATTurbo $\rightarrow$ (DP200, GT450, Scan II, S360)}     \\

& \multicolumn{1}{c}{ScanScope:$[$6$]$}   & \multicolumn{1}{c}{ScanScope $\rightarrow$ (DP200, GT450, Scan II, S360)}                         \\  \midrule

HCC Cell      & \multicolumn{1}{c}{AT2:$[$1,3,6,8,12$]$}   & \multicolumn{1}{c}{AT2 $\rightarrow$ (DP200, GT450, Scan II, S360)}  \\ 

& \multicolumn{1}{c}{ATTurbo:$[$3$]$}  & \multicolumn{1}{c}{ATTurbo $\rightarrow$ (DP200, GT450, Scan II, S360)}     \\

& \multicolumn{1}{c}{ScanScope:$[$3,4,6,8,12$]$}   & \multicolumn{1}{c}{ScanScope $\rightarrow$ (DP200, GT450, Scan II, S360)}   \\
& \multicolumn{1}{c}{GT450:$[$1$]$}   & \multicolumn{1}{c}{GT450 $\rightarrow$ (AT2, DP200, Scan II, S360)}   

\\ 

\midrule
Prostate Cell      & \multicolumn{1}{c}{AT2:$[$1,2,3,5$]$}   & \multicolumn{1}{c}{AT2 $\rightarrow$ (DP200, GT450)}  \\ 

& \multicolumn{1}{c}{GT450:$[$1$]$}  & \multicolumn{1}{c}{GT450 $\rightarrow$ (AT2, DP200)}     \\

& \multicolumn{1}{c}{NanoZoomer:$[$5$]$}   & \multicolumn{1}{c}{NanoZoomer $\rightarrow$ (DP200, GT450)}   \\
& \multicolumn{1}{c}{ UFS:$[$1$]$}   & \multicolumn{1}{c}{UFS $\rightarrow$ (AT2, DP200)}   

\\ \midrule

NASH Cell     & \multicolumn{1}{c}{AT2:$[$1,6$]$}   & \multicolumn{1}{c}{AT2 $\rightarrow$ (DP200, GT450)}  \\ 

& \multicolumn{1}{c}{GT450:$[$1$]$}  & \multicolumn{1}{c}{GT450 $\rightarrow$ (AT2, DP200)}     \\

& \multicolumn{1}{c}{ScanScope:$[$4$]$}   & \multicolumn{1}{c}{ScanScope $\rightarrow$ (DP200, GT450)}   \\

\bottomrule 

\end{tabular}%
}

\end{table*}

\begin{table*}
\caption{Details of synthetic training data created with S-DOTA using SVA in each of the experiments. Note that SVA is not constrained by the one-to-one scanner domain mapping like ST, thus enabling more diverse data to be generated vs. ST. Both original and synthetic data is used to train the pathology models.}
\label{tab:train_data}

\resizebox{\textwidth}{!}{%
\begin{tabular}{c|c|c}
\toprule
Format                                &   \multicolumn{1}{c}{Original Scanner:$[$Labs$]$}    & \multicolumn{1}{c}{Original Scanner $\rightarrow$ (Synthetic Scanners)} \\ \midrule

HCC Tissue      & \multicolumn{1}{c}{AT2:$[$3$]$}   & \multicolumn{1}{c}{AT2 $\rightarrow$ (AT2, ATTurbo, DP200, GT450, NanoZoomer, Scan II, S360)}  \\ 

& \multicolumn{1}{c}{ATTurbo:$[$3,6$]$}  & \multicolumn{1}{c}{ATTurbo $\rightarrow$ (AT2, ATTurbo, DP200, GT450, NanoZoomer, Scan II, S360)}     \\

& \multicolumn{1}{c}{ScanScope:$[$6$]$}   & \multicolumn{1}{c}{ScanScope $\rightarrow$ (AT2, ATTurbo, DP200, GT450, NanoZoomer, Scan II, S360)}                         \\  \midrule

HCC Cell      & \multicolumn{1}{c}{AT2:$[$1,3,6,8,12$]$}   & \multicolumn{1}{c}{AT2 $\rightarrow$ (AT2, ATTurbo, DP200, GT450, NanoZoomer, Scan II, S360)}  \\ 

& \multicolumn{1}{c}{ATTurbo:$[$3$]$}  & \multicolumn{1}{c}{ATTurbo $\rightarrow$ (AT2, ATTurbo, DP200, GT450, NanoZoomer, Scan II, S360)}     \\

& \multicolumn{1}{c}{ScanScope:$[$3,4,6,8,12$]$}   & \multicolumn{1}{c}{ScanScope $\rightarrow$ (AT2, ATTurbo, DP200, GT450, NanoZoomer, Scan II, S360)}   \\
& \multicolumn{1}{c}{GT450:$[$1$]$}   & \multicolumn{1}{c}{GT450 $\rightarrow$ (AT2, ATTurbo, DP200, GT450, NanoZoomer, Scan II, S360)}   

\\ 

\midrule
Prostate Cell      & \multicolumn{1}{c}{AT2:$[$1,2,3,5$]$}   & \multicolumn{1}{c}{AT2 $\rightarrow$ (AT2, ATTurbo, DP200, GT450, NanoZoomer, Scan II, S360)}  \\ 

& \multicolumn{1}{c}{GT450:$[$1$]$}  & \multicolumn{1}{c}{GT450 $\rightarrow$ (AT2, ATTurbo, DP200, GT450, NanoZoomer, Scan II, S360)}     \\

& \multicolumn{1}{c}{NanoZoomer:$[$5$]$}   & \multicolumn{1}{c}{NanoZoomer $\rightarrow$ (AT2, ATTurbo, DP200, GT450, NanoZoomer, Scan II, S360)}   \\
& \multicolumn{1}{c}{ UFS:$[$1$]$}   & \multicolumn{1}{c}{UFS $\rightarrow$ (AT2, ATTurbo, DP200, GT450, NanoZoomer, Scan II, S360)}   

\\ \midrule

NASH Cell     & \multicolumn{1}{c}{AT2:$[$1,6$]$}   & \multicolumn{1}{c}{AT2 $\rightarrow$ (AT2, ATTurbo, DP200, GT450, NanoZoomer, Scan II, S360)}  \\ 

& \multicolumn{1}{c}{GT450:$[$1$]$}  & \multicolumn{1}{c}{GT450 $\rightarrow$ (AT2, ATTurbo, DP200, GT450, NanoZoomer, Scan II, S360)}     \\

& \multicolumn{1}{c}{ScanScope:$[$4$]$}   & \multicolumn{1}{c}{ScanScope $\rightarrow$ (AT2, ATTurbo, DP200, GT450, NanoZoomer, Scan II, S360)}   \\

\bottomrule 

\end{tabular}%
}

\end{table*}
\begin{figure}
\includegraphics[trim=0cm 11cm 0cm 0.5cm, width=\textwidth]{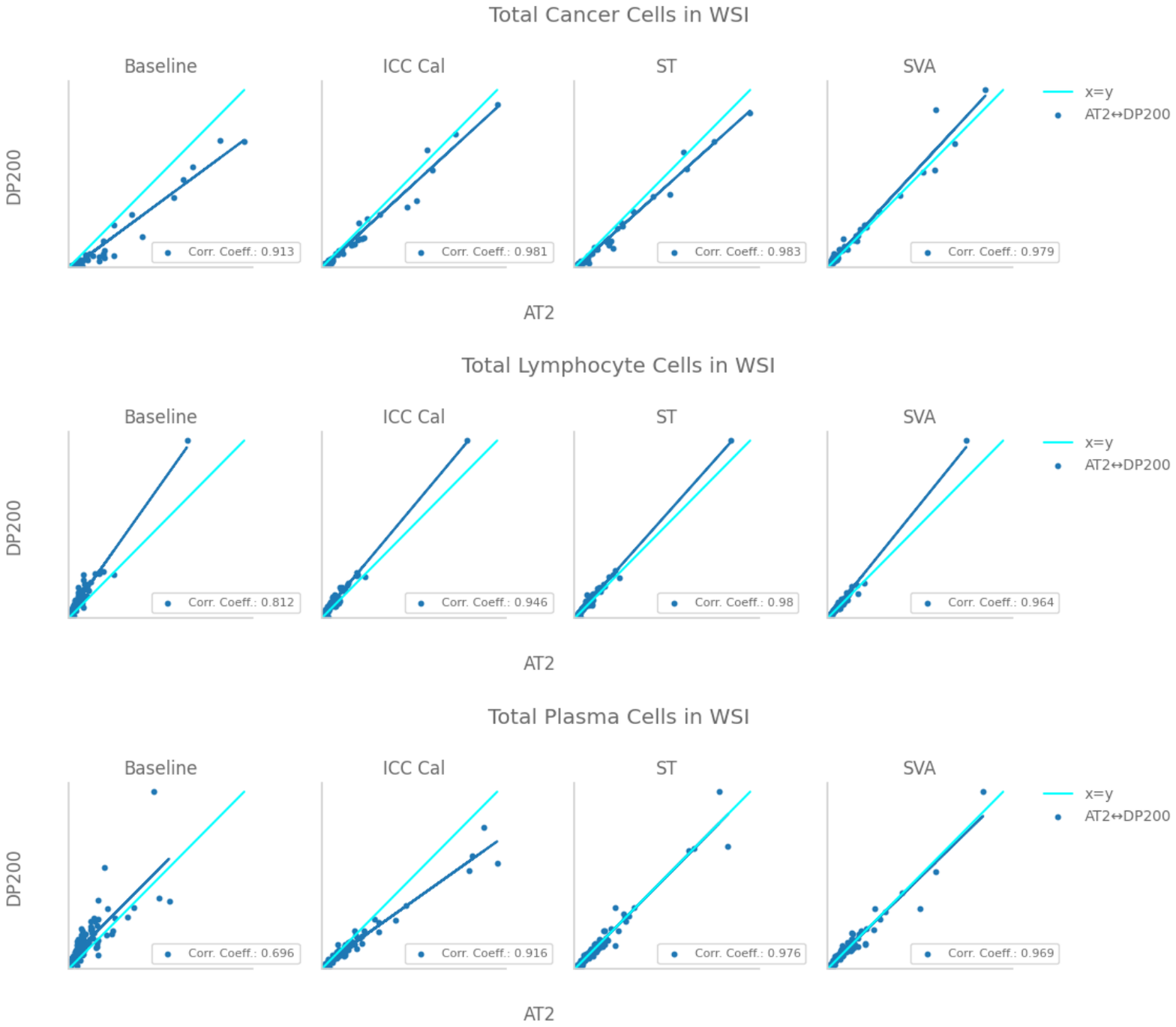}
\caption{\textbf{Prostate Cell} Feature Correlations. Where paired WSIs were available (i.e., the same glass slide scanned on multiple scanners), we inspected the consistency of predicted WSI-wide cell counts. The same glass slide scanned on multiple scanners should result in the same count values, so the closer the points are to y=x, the better. ST, SVA models are closest to y=x, i.e ST, SVA models outperform ICC Cal, Baseline (quantitatively shown by Intra Class Correlation Coefficient).} \label{fig1}
\end{figure}

\end{center}

\end{document}